\documentclass[10pt,twocolumn,letterpaper]{article}

\usepackage[pagenumbers]{wacv} 

\usepackage{graphicx}
\usepackage{amsmath}
\usepackage{amssymb}
\usepackage{booktabs}
\usepackage{multirow}
\usepackage{pifont}
\usepackage{tikz}

\usepackage{color, colortbl}
\definecolor{Gray}{gray}{0.95}

\usepackage[pagebackref,breaklinks,colorlinks]{hyperref}

\usepackage[capitalize]{cleveref}
\crefname{section}{Sec.}{Secs.}
\Crefname{section}{Section}{Sections}
\Crefname{table}{Table}{Tables}
\crefname{table}{Tab.}{Tabs.}

\def\wacvPaperID{1171} 
\def\confName{WACV}
\def\confYear{2025}

\newcommand{\tikzxmark}{%
\tikz[scale=0.23] {
    \draw[line width=0.7,line cap=round] (0,0) to [bend left=6] (1,1);
    \draw[line width=0.7,line cap=round] (0.2,0.95) to [bend right=3] (0.8,0.05);
}}

\begin{document}

\title{COSNet: A Novel Semantic  Segmentation Network using Enhanced Boundaries in Cluttered Scenes}

\author{
Muhammad Ali$^1$ \quad Mamoona Javaid$^2$ \quad Mubashir Noman$^1$ \quad Mustansar Fiaz$^3$ \quad Salman Khan$^1$
\vspace{0.4em}
\\
$^1$Mohamed bin Zayed University of AI, U.A.E \quad $^2$Institute of Space Technology, Pakistan  \\ $^3$IBM Research
}

\maketitle

\begin{abstract}
Automated waste recycling aims to efficiently separate the recyclable objects from the waste by employing vision-based systems. However, the presence of varying shaped objects having different material types makes it a challenging problem, especially in cluttered environments. Existing segmentation methods perform reasonably on many semantic segmentation datasets by employing multi-contextual representations, however, their performance is degraded when utilized for waste object segmentation in cluttered scenarios.
In addition, plastic objects further increase the complexity of the problem due to their translucent nature.
To address these limitations, we introduce an efficacious segmentation network, named COSNet, that uses boundary cues along with multi-contextual information to accurately segment the objects in cluttered scenes. 
COSNet introduces novel components including feature sharpening block (FSB) and boundary enhancement module (BEM) for enhancing the features and highlighting the boundary information of irregular waste objects in cluttered environment. 
Extensive experiments on three challenging datasets including \textit{ZeroWaste-f} \cite{bashkirova2022_zerowaste}, \textit{SpectralWaste} \cite{casao2024spectralwaste}, and \textit{ADE20K} \cite{Zhou2017ScenePT_ade20k} demonstrate the effectiveness of the proposed method. Our COSNet achieves a significant gain of 1.8\% on ZeroWaste-f and 2.1\% on SpectralWaste datasets respectively in terms of mIoU metric. Source code is available at \url{https://github.com/techmn/cosnet}.
\end{abstract}

\section{Introduction}
\label{sec:intro}
With the upsurge of world population and urbanization, the production of waste is rising which may encounter detrimental effects on the environment \cite{waste_2018}. The effective recycling of the waste objects such as plastic, metal, paper, etc., is therefore a critical task and requires careful handling procedures \cite{bashkirova2022_zerowaste}. Although material recycling facilities (MRFs) or plants that separate and prepare recyclable materials utilize heavy machinery for recycling of waste material. However, the manual labor may risk their health by getting exposed to the bare metal and unhygienic medical waste \cite{bashkirova2022_zerowaste, GUNDUPALLI201756}. It is therefore necessary to develop an automated system that can effectually separate the recyclable objects from the waste material.
Furthermore, the translucent nature of the plastic objects and the presence of a cluttered background make the automated separation of the recyclable waste challenging. 

In literature, various waste classification methods have been proposed \cite{xia2024yolo, mao2021recycling, feng2022intelligent, meng2022mobilenet}, they pose limitations in the presence of complex scenarios where there exists an unclear boundary information.
Recently, Bashkirova et. al. \cite{bashkirova2022_zerowaste} and Casao et al.  \cite{casao2024spectralwaste} introduced challenging datasets, called  ZeroWaste and SpectralWaste, for the task of waste object segmentation in a cluttered scenes, respectively. 
Current deep learning approaches \cite{chen2018_deeplabv3_plus, badrinarayanan2016_segnet, liu2021swin, yuan2021segmentation_transformer, Zhao_2017_pspnet, chen2017_deeplabv3} have shown promising results on semantic segmentation task by utilizing the multi-scale contextual features in an encoder-decoder architecture.
For instance, SegNet \cite{badrinarayanan2016_segnet} introduces the encoder-decoder framework to benefit from both the fine grained boundary information of shallow layers as well as the rich semantic features of last layers, DeepLabv3 \cite{chen2017_deeplabv3} exploits the atrous convolutions to capture multi-scale information for better segmentation performance, PSPNet \cite{Zhao_2017_pspnet} uses the pyramid pooling module to capture the multi-contextual representations, and DeepLabv3+ \cite{chen2018_deeplabv3_plus} further optimizes the method by using encoder-decoder framework and depth-wise separable convolutions. Furthermore, Xie et. al. \cite{xie2020_segment_transparent} discuss the importance of boundary cues for accurate segmentation of transparent objects.
Despite of utilizing the encoder-decoder architecture \cite{badrinarayanan2016_segnet, chen2017_deeplabv3} for capturing rich contextual information by pooling features and exploiting the fine detail information from shallow layers, the existing well-proven methods such as \cite{chen2018_deeplabv3_plus} still strive to accurately segment the waste objects in cluttered scenes. 
Conversely, transformer-based methods \cite{yuan2021segmentation_transformer, liu2021swin} have shown remarkable performance for segmentation tasks due to their ability to capture global contextual relationships. However, these methods rely on a large volume of annotated data which can be expensive and may require strenuous efforts especially for specialized applications \cite{bashkirova2022_zerowaste}. Additionally, the dominance of global representations may limit the ability of transformers to focus on fine details that are necessary to segment the translucent and deformable-shaped objects in cluttered scenes. 
Therefore, an explicit mechanism is required to highlight the boundaries of such objects for better segmentation results.


To address the limitations of the aforementioned methods, we propose an effectual segmentation network, called COSNet 
(Cluttered Objects’ semantic Segmentation Network)
, that highlights the boundary cues and utilizes the multi-scale information in an encoder-decoder architecture for better segmentation of waste objects in cluttered scenes. Inspired from the unsharp masking \cite{MLSNA2009495_unsharp_masking}, we propose a feature sharpening block (FSB) that benefits from both enlarged receptive field and implicit feature boosting techniques. 
Specifically, we capture the multi-scale contextual representations using atrous convolutions to enlarge the receptive field and enhance the boundary details by exploiting an implicit sharpening module within our FSB module.  Furthermore, we complement our COSNet by exploring explicit high-boost filtering. To achieve this, we propose a boundary enhancement module (BEM), as an intermediate network,  to supplement the COSNet in obtaining finer feature representations.
To summarize, our contributions are as below:

\begin{itemize}
\item We propose an effective backbone network that effectively utilizes the multi-scale feature representations and highlights the boundary information to accurately segment the waste material such as plastic objects in a cluttered background.
\item We propose BEM that is utilized in an encoder-decoder segmentation framework for further boosting the feature representations obtained from the backbone network.
\item Extensive experimentation on three challenging semantic segmentation datasets (ZeroWaste \cite{bashkirova2022_zerowaste}, SpectralWaste \cite{casao2024spectralwaste} and ADE20K \cite{Zhou2017ScenePT_ade20k}) reveal the merits of the proposed contributions.
\end{itemize}

\section{Related Work}
\subsection{Waste Detection and Segmentation}
Waste detection has recently gained immense popularity due to various challenges including cluttered backgrounds,  translucent objects, illumination variations, and objects' varying shapes and sizes. 
Since the advent of deep learning, various convolutional neural network (CNNs) based methods have been introduced for waste sorting \cite{qin2022robust, xia2024yolo, mao2021recycling, feng2022intelligent, meng2022mobilenet, tian2024garbage, hossen2024reliable}. 
Qin et al. \cite{qin2022robust} proposed a garbage classification framework by introducing a small amount of background and noise to alleviate the overfitting and improve the robustness.
Xia et al. \cite{xia2024yolo} exploit MobileViTv3 \cite{wadekar2022mobilevitv3} features to capture the global and local features for garbage detection. 
Mao et al. \cite{mao2021recycling} utilize DenseNet121 \cite{huang2017densely} to introduce a genetic algorithm that fine-tunes the parameters of the fully connected layers of the model.
Feng et al. \cite{feng2022intelligent} used the EfificientNet \cite{koonce2021efficientnet} as a backbone feature extractor and introduced the efficient channel attention (ECA) and coordinate attention (CA) modules to capture the waste objects.
Meng et al. \cite{meng2022mobilenet} employ FPN along with a focal loss to reduce the imbalance between foreground and background samples to enhance the detector effect. 
Tian et al. \cite{tian2024garbage} exploit the MobileNetV3 \cite{howard2019searching} and introduce the CBAM \cite{woo2018cbam} to enhance the spatial features for garbage recognition. 
Hossen et al. \cite{hossen2024reliable} introduce a recyclable waste classification (RWC-Net) to classify the wastes. 

Though these above-mentioned waste classification methods demonstrate great performance, these approaches present limitations in various scenarios including a wide range of waste types and unstructured backgrounds.
Various efforts have been made to segment the waste from the background \cite{bashkirova2022_zerowaste, qi2024nuni, Ali2024FANet, yudin2024hierarchical, corrigan2023real, casao2024spectralwaste, cascina2023resource, rahman2024kitchen}.
Bashkirova et al. \cite{bashkirova2022_zerowaste} utilizes DeepLabv3+ as backbone features and exploits atrous spatial pyramid pooling (ASPP) to encode the multi-scale semantic features for segmentation maps.
Qi et al. \cite{qi2024nuni} uses non-uniform data augmentation tailored to real waste classification scenarios.  
Yudin et al. \cite{yudin2024hierarchical} propose a hierarchical neural network, called H-YC, that can perform waste classification and segmentation.
Corrigan et al. \cite{corrigan2023real} propose a method to detect underwater litter using Mask RCNN \cite{he2017mask}. 
Casao et al. \cite{casao2024spectralwaste}  benefits from RGB and HSI modalities for waste object segmentation. 
Cascina et al. \cite{cascina2023resource} utilize different architectures for constrained semantic segmentation models for segmenting recyclable waste.
Rahman et al. \cite{rahman2024kitchen} employ generalized semantic
segmentation techniques over the kitchen food waste. Ali et al. \cite{Ali2024FANet} introduce a feature amplification network (FANet) that is capable of capturing the semantic information at multi-stages using a feature enhancement module. 
Existing methods still struggle to localize the objects due to cluttered backgrounds and unclear boundaries between the waste objects and backgrounds. 


\subsection{Semantic Segmentation}
Recently, deep learning techniques have shown promising performance over semantic segmentation tasks \cite{chen2018_deeplabv3_plus, liu2021swin, yuan2021segmentation_transformer, lin2023scale, mininet2020, hou2024conv2former, wang2023internimage, neurips2021_segformer, liu2022convnext, yang2022focalnet}.
Chen at al. \cite{chen2018_deeplabv3_plus} extended the DeepLabv3 \cite{chen2017_deeplabv3} by utilizing depthwise separable convolution to both Atrous Spatial Pyramid Pooling and decoder modules. 
MiniNet-v2 \cite{mininet2020} is built on ConvNext for efficient semantic segmentation for real-time robotic applications.
Wang et al. \cite{wang2023internimage} introduce the deformable convolution as the core operator to enhance the receptive field for semantic segmentation.
Lin et al. \cite{lin2023scale} propose a multi-head mixed convolution (MHMC) operation along with a scale-aware aggregation (SAA) module to exploit multi-scale context features.
Hou et al. \cite{hou2024conv2former} leverage the convolutional modulation by utilizing a large kernel to enlarge the receptive field. 
Xie et al. \cite{neurips2021_segformer} propose Segformer which exploits transformers with lightweight multilayer perception (MLP) decoders combining both local attention and global attention to render powerful representations.
Liu et al. \cite{liu2022convnext} reexamine the design spaces to test the limits of what a pure ConvNet is and propose ConvNext.
Yang et al. \cite{yang2022focalnet} propose a FocalNet that modulates the features at different granularities to exploit features from multiple scales. 
Fan et al. \cite{Fan_2021_CVPR} redesigned the BiSeNet \cite{Yu_ECCV_BiSeNet_2018} and propose a detail aggregation module that utilizes the static Laplacian filter to highlight the boundaries of the objects for better segmentation performance. 
Similarly, Kim et al. \cite{kim2020devilboundaryexploitingboundary} introduce B2Inst model for instance segmentation task that uses static Laplacian kernel to extract the boundaries of the predicted and ground truth masks. Afterwards, boundary loss is utilized to learn the boundaries of the object instances.

In contrast, we propose a multi-contextual features
extraction and sharpening module that strives to capture implicit local information and sharpening information to handle complex shapes and sizes of waste objects. Furthermore, our explicit boundary enhancement module learns salient features to handle the background clutter and delineates the translucent waste objects from backgrounds.

\section{Background}
In image processing, unsharp masking \cite{MLSNA2009495_unsharp_masking} is a common technique to enhance the edges and details of the objects in an image. The unsharp masking is a filter that emphasizes the high-frequency content.
In the unsharp masking technique, a blurred (unsharp) version of the image is generated and subtracted from the original input image. The output, called the mask, focuses on the edges and fine details of the image. Later the mask is added to the original image which results in a sharpened image emphasizing the texture and details of the image.

Let $f(x,y)$ and $ \bar f(x,y)$ be the input and its blurred version in latent space, and the mask $ g_{mask}(x,y)$ can be obtained using equations as:
\begin{equation}
\label{eq:sharp}
    g_{mask}(x,y) = f(x,y) -  \bar f(x,y).
\end{equation}
Finally, the sharpened feature map $g(x,y)$ can be obtained follows as:
\begin{equation}
\label{eq:sharp2}
    g(x,y) = f(x,y) +  k * g_{mask}(x,y),
\end{equation}
where $k \in (k \geq 0)$ represents the weight. When $k==1$ we have
unsharp masking and when $k>1$ the process is referred to as high boost filtering. Motivated by this, we leverage our sharpening module (SM) and boundary enhancement module (BEM) as described in Sec. \ref{sssec:fsb} and \ref{sssec:bem}, respectively. 

\section{Method}
In this section, we describe the proposed approach in detail.

\begin{figure}[t]
\centering
 \includegraphics[width=1.0\linewidth]{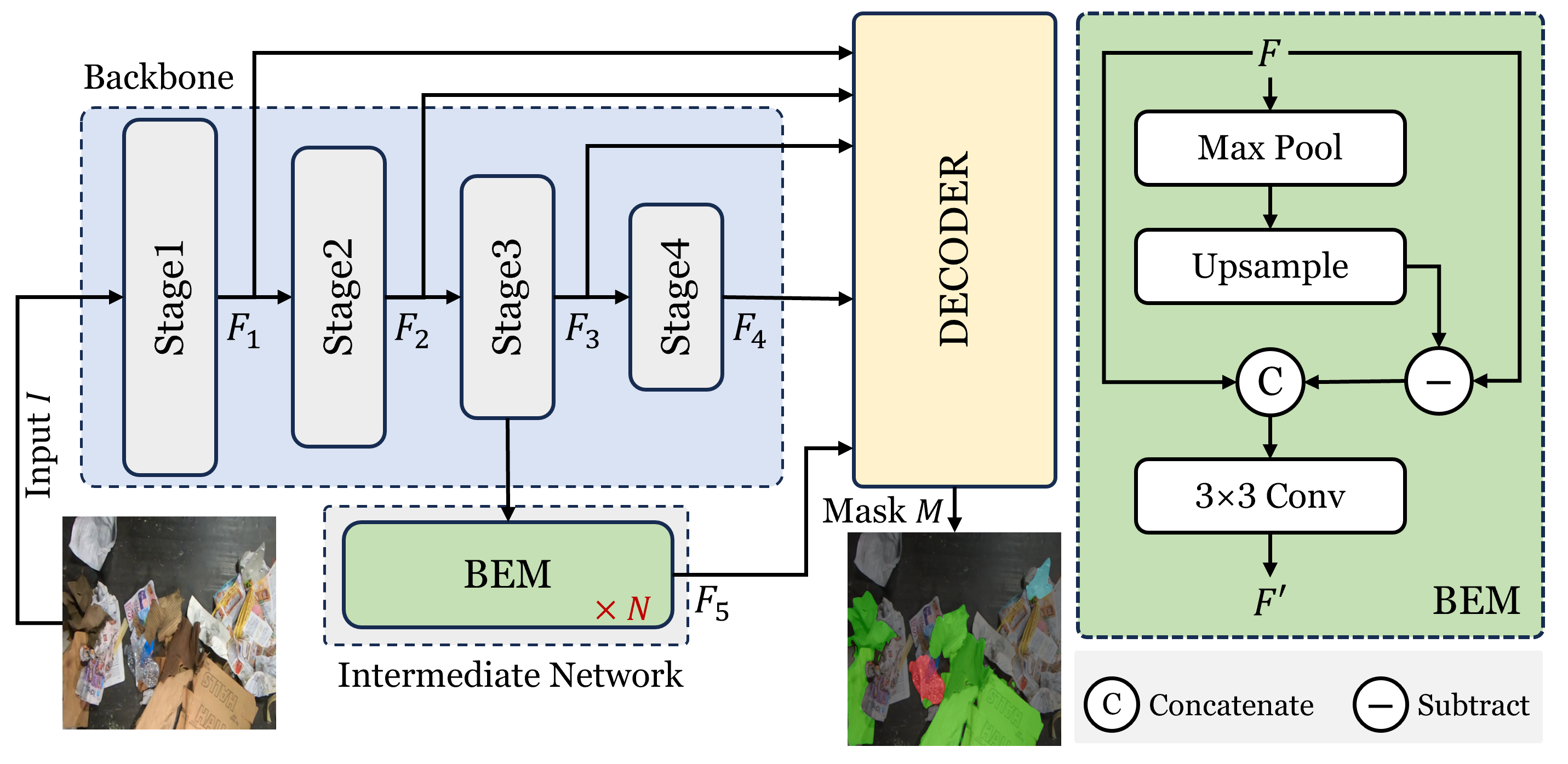} 
\caption{Illustration of the overall architecture of the proposed segmentation framework named COSNet. The proposed framework utilizes enhanced backbone network to obtain rich multi-scale representations through FSBs (sec.~\ref{sssec:fsb}) of the backbone network. The intermediate network further enhances the boundary details of the third-stage features by means of a boundary enhancement module (BEM). Finally, the multi-scale feature maps, $F_i$ where $i \in {1,2,3,4,5}$, are passed to the decoder to obtain the segmentation mask $M$. }
 \label{fig:main_framework}
\end{figure}

\subsection{Overall Architecture}
In Fig. \ref{fig:main_framework}, we illustrate the overall architecture of the proposed segmentation framework named COSNet. The proposed framework consists of a backbone network, an intermediate network, and a decoder. The backbone network consists of four stages and utilizes FSBs to provide the multi-scale feature representations. The intermediate network uses the BEM to further enrich the feature maps of the third stage.
The utilization of intermediate network is task-relevant such as further enhancement of object boundaries in the case of translucent objects and may be replaced by the identity module.
Finally, the feature representations from the backbone and intermediate networks are fed to a decoder that effectively fuses the multi-scale feature representations to obtain segmentation masks.

\begin{figure*}[t]
\centering
 \includegraphics[width=1.0\linewidth]{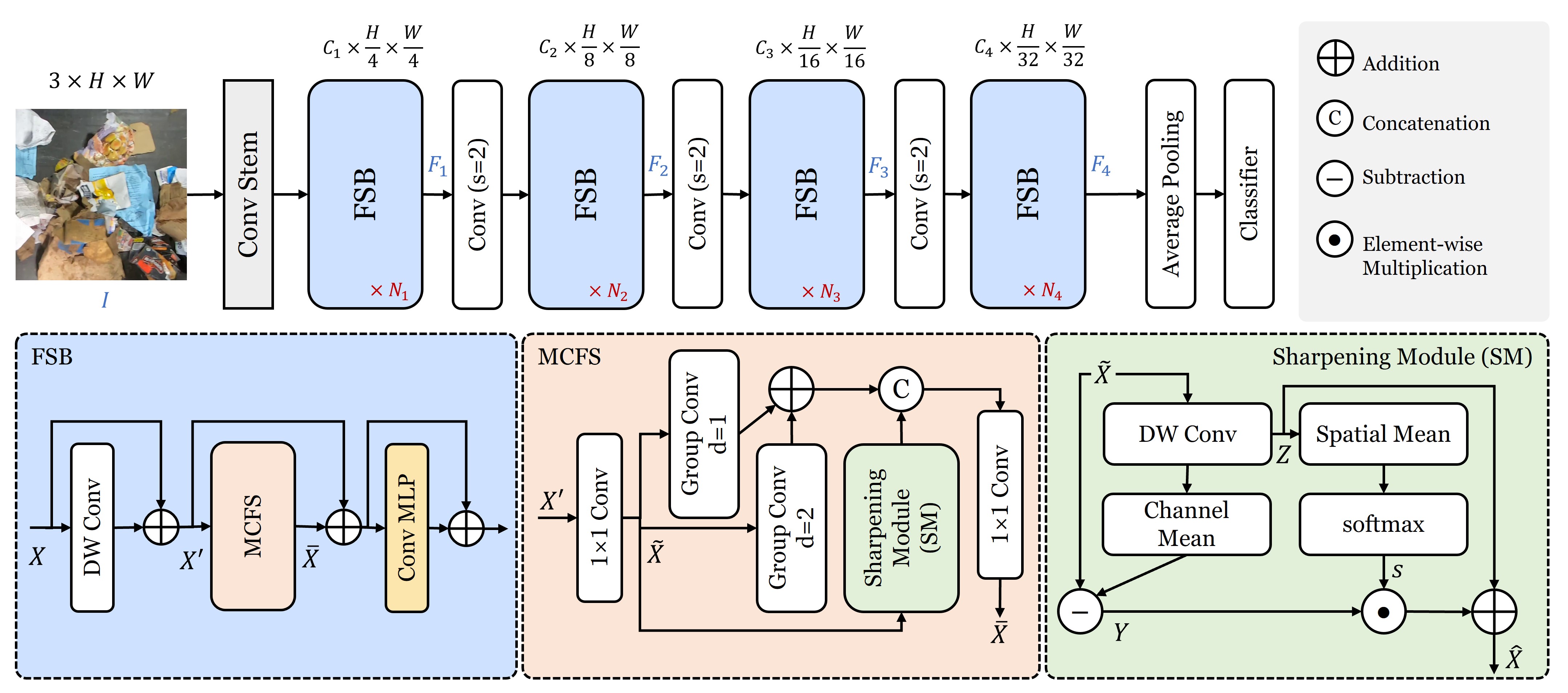} 
\caption{Illustration of the backbone network of the proposed COSNet. The backbone network extracts multi-scale features at four scale levels by utilizing feature sharpening blocks (FSBs). The core component of FSB is multi-contextual features extraction and sharpening (MCFS) module that uses dilated convolutions to extract multi-contextual representations and a learnable sharpening module to enhance the boundary information. }
 \label{fig:backbone_fsb}
\end{figure*}
\subsection{COSNet Backbone}
The backbone network comprises of four stages to obtain the multi-scale feature representations $F_i$ where $i \in {1,2,3,4}$ as shown in Fig. \ref{fig:backbone_fsb}. Each stage utilizes a series of feature sharpening blocks (FSBs) to obtain rich feature representations.
The COSNet backbone takes an input image $I$ and feeds it to a convolution stem layer that projects the image to feature space and downsamples the spatial resolution by four times. 
The last three stages use a convolution layer with a stride value of two to reduce the spatial resolution of the feature maps before passing them to feature sharpening blocks.
\subsubsection{Feature Sharpening Block (FSB)}
\label{sssec:fsb}
The design of the FSB is illustrated in Fig. \ref{fig:backbone_fsb}. The FSB is composed of a depth-wise convolution layer, a multi-contextual feature sharpening module (MCFS) and a convolutional multi-layer perceptron (MLP). Being a key module of the FSB, MCFS is responsible for capturing the multi-contextual information while focusing on the object boundaries for obtaining the better segmentation maps. Inspired by \cite{chen2017_deeplabv3}, we utilize atrous convolutions to capture multi-contextual features. Although multi-contextual information plays a significant role in better segmentation performance \cite{chen2018_deeplabv3_plus, chen2017_deeplabv3}, however, it struggles to highlight boundary cues for translucent and cluttered objects. For this reason, we introduce an implicit feature sharpening module (SM) that highlights the boundary information of objects in cluttered scenarios.

Given feature maps $X^{'}$, MCFS first projects the features by using a $1 \times 1$ convolution layer to obtain features $\Tilde{X} \in \mathbb{R}^{\Tilde{C} \times \Tilde{H} \times \Tilde{W}}$. Then, we utilize two group convolution layers having different dilation values to extract multi-contextual features. In parallel, feature maps $\Tilde{X}$ are passed to sharpening module (SM) to get the boundary enhanced features. Lastly, we concatenate the multi-contextual features and the boundary enhanced features along channel dimension and utilize a $1 \times 1$ convolution layer to get enhanced feature representations.

The sharpening module (SM) first utilizes a depth-wise convolution layer to encode local representations and separately enhance the structural information in each channel and obtain features $Z$. Afterwards, we compute mean of the feature maps ($Z$) along channel dimension to obtain the averaged response of feature maps. To highlight the fine details, we subtract the averaged response from input features $\Tilde{X}$ to get features $Y$. Additionally, we compute the spatial mean of each channel of features $Z$ and apply softmax operation to obtain a sharpening factor ($s \in \mathbb{R}^{\Tilde{C} \times 1 \times 1}$) for each channel of feature maps $Y$. We multiply the sharpening factors $s$ with the features $Y$ and add it to the $Z$ to obtain the sharpened features $\hat{X}$.

\subsection{Boundary Enhancement Module (BEM)}
\label{sssec:bem}
Our explicit boundary enhancement module is utilized to further enhance the feature maps of the third stage of the backbone network as shown in Fig.~\ref{fig:main_framework}. The third stage features corresponds to high level semantics as well as preserves the fine details of the regions. The idea is to focus on the high frequency details of the feature maps similar to the process of highlighting the boundaries by subtracting the blurred image from the original one \cite{MLSNA2009495_unsharp_masking}. BEM takes input feature maps $F$ and utilize max pooling operation to extract the significant features. The pooled features have smaller spatial resolution compared to the input features $F$. Therefore, pooled features are upsampled back to the input spatial resolution. To highlight the fine details, we perform the subtraction operation between the input $F$ and upsampled features. The highlighted features are concatenated with input features $F$ and a $3 \times 3$ convolution layer is utilized to obtain the enhanced feature maps $F^{'}$.  

\begin{table}[t]
\centering
\caption{Comparison of our framework with state-of-the-art methods for semantic segmentation on the test set of the ZeroWaste-f dataset. We report the results in terms of mIoU and pixel accuracy. The best results are highlighted in bold text.}
\vspace{-0.5em}
\setlength{\tabcolsep}{9pt}
\begin{tabular}{l|c|c}
\hline
\rowcolor{Gray}

Method & mIoU (\%)$\uparrow$ & Pix. Acc. (\%)$\uparrow$ \\
\hline
CCT \cite{9157032_cct} &  29.32 & 85.91 \\
ReCo \cite{ReCo} & 52.28  & 89.33 \\
DeepLabv3+ \cite{chen2018_deeplabv3_plus} & 52.13 &  91.38 \\
FANet \cite{Ali2024FANet} & 54.89 & 91.41 \\
InternImage-T \cite{interimage_cvpr_2023} & 54.33 & 91.84 \\
\textbf{COSNet (Ours)} &  \textbf{56.67} & \textbf{91.91} \\
\hline
\end{tabular}

\label{tab:comparison_of_cosnet_on_zerowaste}
\end{table}

\begin{table}[t]
\centering
\caption{Class-wise IoU (\%) score of our COSNet framework with DeepLabv3+ on the test set of the ZeroWaste-f dataset. The proposed COSNet achieves significant gains in IoU score for all classes. The best results are highlighted in bold text.}
\vspace{-0.5em}
\setlength{\tabcolsep}{18pt}
\begin{tabular}{l|c|c}
\hline
\rowcolor{Gray}
 & DeepLabv3+ & Ours \\
\hline
Background & 91.02 & \textbf{91.44} \\
Cardboard & 54.47  & \textbf{59.13} \\
Soft Plastic & 63.18 & \textbf{65.92} \\
Rigid Plastic & 24.82 & \textbf{37.24}  \\
Metal & 27.14  & \textbf{29.61} \\
\hline
\end{tabular}
\vspace{-0.5em}

\label{tab:classwise_comparison_on_zerowaste}
\end{table}

\section{Experiments}
Here we discuss the datasets, implementation details,  quantitative, and qualitative comparison of our method with state-of-the-art methods.
\subsection{Datasets and Evaluation Metrics} To evaluate the performance of proposed framework on cluttered scenes, we utilize ZeroWaste-f \cite{bashkirova2022_zerowaste} and SpectralWaste \cite{casao2024spectralwaste} datasets. Additionally, we report the performance of our framework on a well known public semantic segmentation dataset ADE20K \cite{Zhou2017ScenePT_ade20k}.

\begin{figure*}[t!]
\centering
 \includegraphics[width=1.0\linewidth]{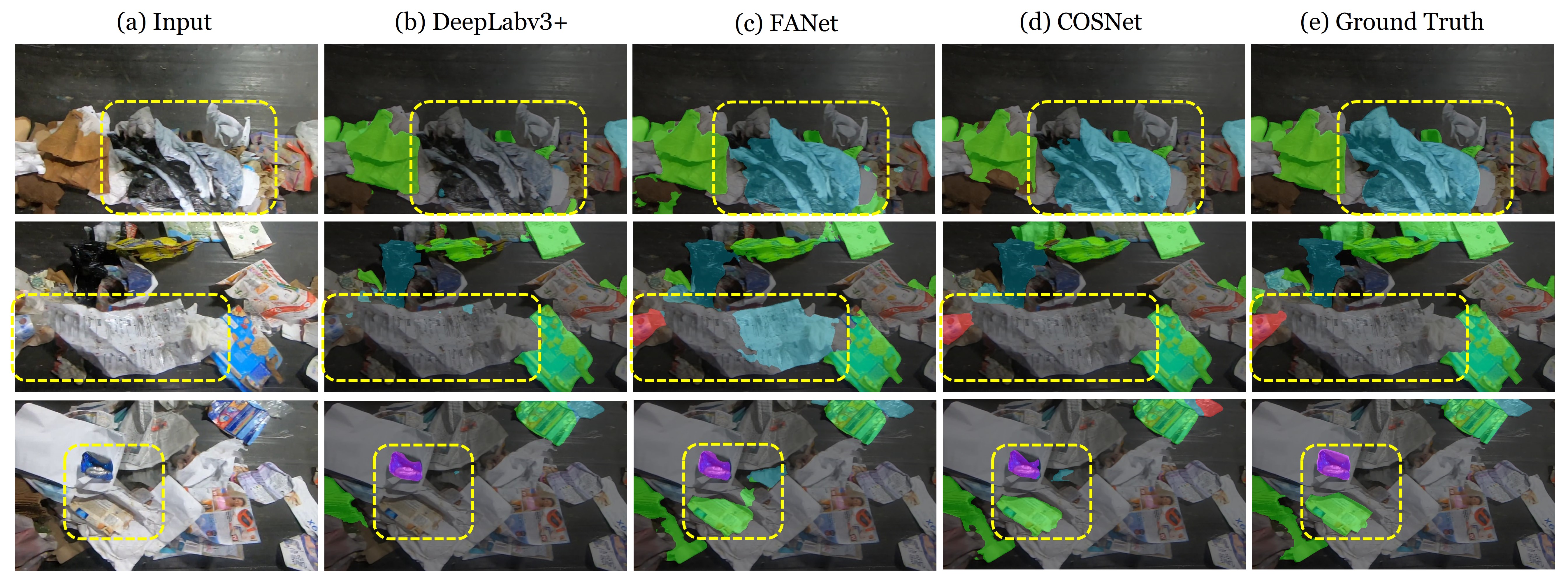} 
\caption{Here we present the segmentation results of our COSNet on ZeroWaste-f \cite{bashkirova2022_zerowaste} dataset. Our method can accurately segment the different waste material types compared to the DeepLabv3+ \cite{chen2018_deeplabv3_plus} and recently introduced FANet \cite{Ali2024FANet} as highlighted in yellow boxes. }
 \label{fig:zero_waste_results}
\end{figure*}
\begin{table*}[t!]
\centering
\caption{Comparison of our framework with state-of-the-art methods for semantic segmentation on the test set of the SpectralWaste \cite{casao2024spectralwaste} dataset. We report the results in terms of mIoU and individual class IoUs. The best results are highlighted in bold text. The $*$ represents that the reported results are from SpectralWaste \cite{casao2024spectralwaste} having smaller backbone networks.}
\setlength{\tabcolsep}{9pt}
\begin{tabular}{l|c|cccccc}
\hline
\rowcolor{Gray}
\multicolumn{1}{l|}{\multirow[t]{2}{*}{Method}} & \multicolumn{1}{|c|}{\multirow[t]{2}{*}{mIoU $(\%)\uparrow$}} & \multicolumn{6}{c}{IoU per Class $(\%)\uparrow$} \\
\cline{3-8}
\rowcolor{Gray}
 &  & Film & Basket & Cardboard & Video Tape & Filament & Trash Bag \\
\hline
MiniNet-v2$^*$ \cite{mininet2020} & 44.5 & 63.1 & 58.9 & 55.4 & 30.6 & 10.0 & 49.2 \\
SegFormer-B0$^*$ \cite{neurips2021_segformer} & 48.4 & 66.9 & 71.3 & 48.9 & 33.6 & 15.2 & 54.6 \\
\hline
FANet \cite{Ali2024FANet} & 67.83 & 72.47 & 82.98 & \textbf{75.26} & 41.28 & 67.65 & 67.36 \\
InternImage-T \cite{interimage_cvpr_2023} & 47.99 & 42.38 & 82.8 & 69.1 & 41.39 & 16.5 & 35.77 \\
\textbf{COSNet (ours)} & \textbf{69.96} & \textbf{77.61} & \textbf{83.65} & 75.14 & \textbf{42.95} & \textbf{69.06} & \textbf{71.38} \\
\hline
\end{tabular}
\label{tab:comparison_of_cosnet_on_spectralwaste}
\end{table*}

\textit{ZeroWaste-f} \cite{bashkirova2022_zerowaste} is a challenging waste segmentation dataset that contains object classes of four material types including cardboard, soft plastic, rigid plastic, and metal. The dataset comprises of 3002, 572, and 929 images in train, validation, and test sets respectively.

\textit{SpectralWaste} \cite{casao2024spectralwaste} is another recently introduced dataset which has a cluttered background and translucent waste objects. It is collected from a waste processing facility and contains synchronized RGB and hyperspectral images. The dataset comprises of six object classes containing cardboard, film, basket, filaments, video tape, and trash bags. The dataset is challenging due to the presence of clutter and brightness variations. SpectralWaste dataset contains 514, 167, and 171 images in train, validation, and test sets respectively. In our experiments, we only utilize the RGB images of the dataset.

\textit{ADE20K} \cite{Zhou2017ScenePT_ade20k} is a well known public semantic segmentation dataset comprising of 150 object classes in diverse scenarios. The dataset contains 20210 training and 2000 validation images.

For performance evaluation, we report the \textit{mean intersection over union} (mIoU) and \textit{pixel accuracy} score on the respective datasets.

\begin{figure*}[t!]
\centering
 \includegraphics[width=0.9\linewidth]{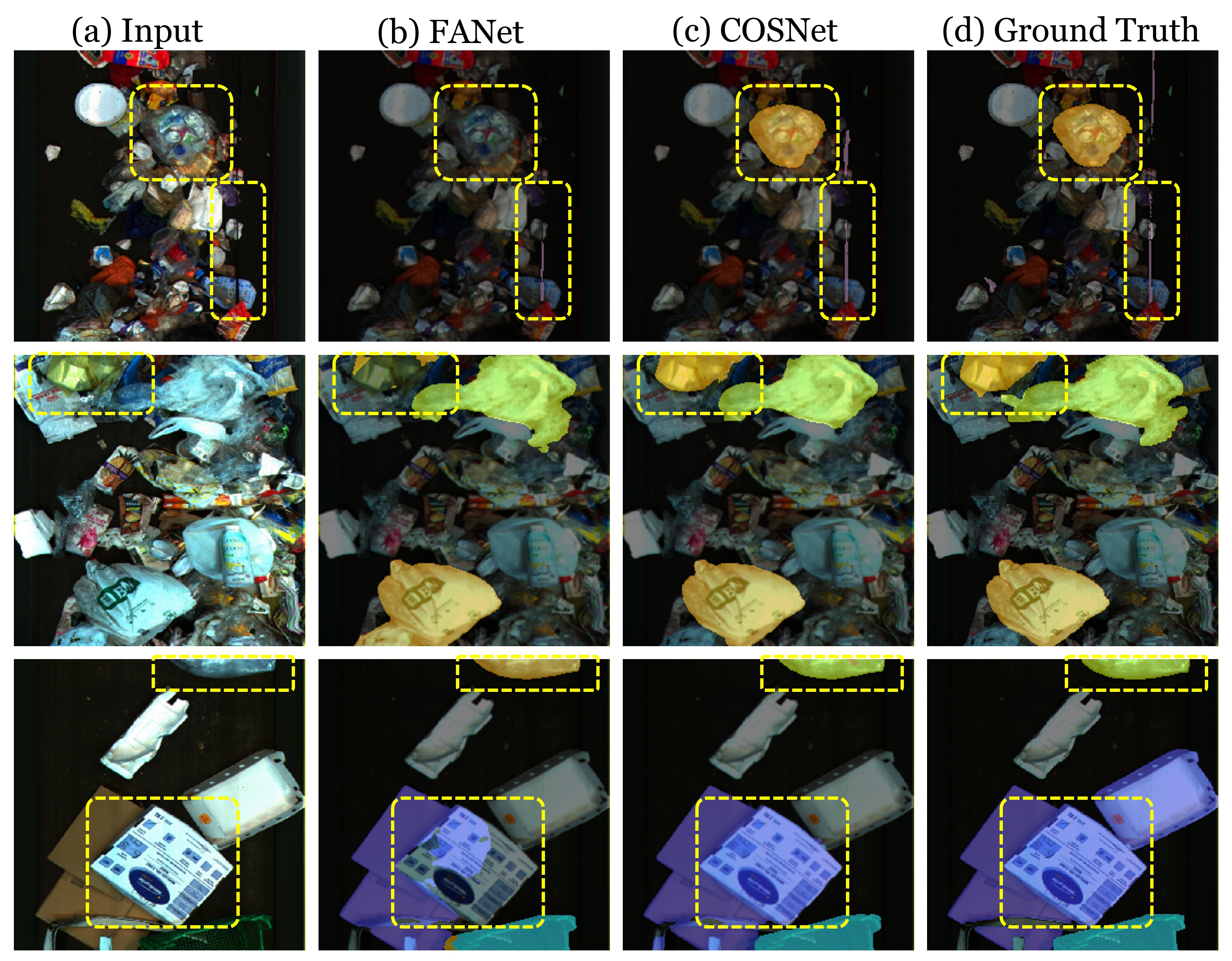} 
\caption{Comparison of the segmentation results of proposed COSNet on Spectral Waste (RGB images) dataset. The proposed COSNet can accurately segment the different waste material types compared to the FANet as highlighted in yellow boxes.}
 \label{fig:spectral_waste_results}
 \vspace{-0.5em}
\end{figure*}

\subsection{Implementation Details}
The proposed method is implemented in Pytorch by utilizing MMSegmentation library \cite{mmseg2020}. We utilize NVIDIA RTX A6000 GPU for experimentation. In training phase, we initialize the backbone with ImageNet \cite{imagenet_cvpr09} pretrained weights and optimize the model by using AdamW \cite{loshchilov2018_adamw} optimizer with initial learning rate of 9e-5. The learning rate is decayed by utilizing  polynomial scheduler. We train the model for 40k iterations for the ZeroWaste-f \cite{bashkirova2022_zerowaste} and SpectralWaste \cite{casao2024spectralwaste} datasets, and 160k iterations for ADE20K \cite{Zhou2017ScenePT_ade20k} dataset. The shorter side of images are randomly resized to 512 pixels and we utilize random crop of 512 $\times$ 512 pixels during training.
We train the COSNet backbone on ImageNet1k \cite{imagenet_cvpr09} for 300 epochs. For each segmentation experiment, we load the ImageNet1k pre-trained weights in the backbone and utilize the UperNet \cite{xiao2018_upernet} decoder to obtain a segmentation mask.

\subsection{Quantitative Results}
\subsubsection{ZeroWaste-f \cite{bashkirova2022_zerowaste}}
We report the semantic segmentation performance of COSNet on ZeroWaste-f dataset in Tab.~\ref{tab:comparison_of_cosnet_on_zerowaste}. Among existing methods, DeepLabv3+ \cite{chen2018_deeplabv3_plus} provides promising results by achieving the mIoU score of 52.13\% while FANet \cite{Ali2024FANet} achieves  the mIoU score of 54.89\%. However, our proposed COSNet performs favorably and sets new state-of-the-art score of 56.67\% in terms of mIoU metric.

We further compare the class-wise IoU score of the existing best DeepLabv3+ (as reported in \cite{bashkirova2022_zerowaste}) with our COSNet in Tab.~\ref{tab:classwise_comparison_on_zerowaste}. We notice that there is a significant improvement in IoU for all class objects especially rigid plastic. The reason is that COSNet enhances the boundaries of the objects that supplement the model's performance to better discriminate the objects in cluttered scenes.

\subsubsection{SpectralWaste \cite{casao2024spectralwaste}}
Tab.~\ref{tab:comparison_of_cosnet_on_spectralwaste} compare the performance of COSNet with the existing state-of-the-art waste segmentation methods as well as the methods reported in \cite{casao2024spectralwaste}. First two rows of the table demonstrate the performance of the networks that utilize shallower backbones as reported in \cite{casao2024spectralwaste}. For fair comparison, we compare the results of COSNet with the recent waste segmentation method FANet \cite{Ali2024FANet}. As illustrated from the table, proposed COSNet achieves a significant gain of 2.1\% in terms of mIoU metric. In case of class-wise comparison, FANet \cite{Ali2024FANet} has slightly better IoU score of 75.3\% for \textit{Cardboard} objects compared to 75.1\%. However, COSNet surpasses the FANet \cite{Ali2024FANet} for the other five classes by obtaining significant gains of approximately 4.02\% and 5.14\% for \textit{Trash Bag} and \textit{Film} object classes, respectively.

\subsubsection{ADE20K \cite{Zhou2017ScenePT_ade20k}}
Tab.~\ref{tab:comparison_of_cosnet_on_ade20k} presents the comparison of our COSNet with the existing well-established backbones networks on ADE20K \cite{Zhou2017ScenePT_ade20k} dataset. Despite having \textit{less number of parameters}, our COSNet outperforms the existing methods and achieves a gain of 0.3\% in terms of mIoU metric.
We further compare the average latency on the test set of ADE20k dataset using crop size of $512 \times 512$ on Nvidia A100 GPU.
These results further validate the effectiveness of FSB for semantic segmentation task.

\subsection{Qualitative Results}
\begin{figure}[t]
\centering
 \includegraphics[width=1.0\linewidth]{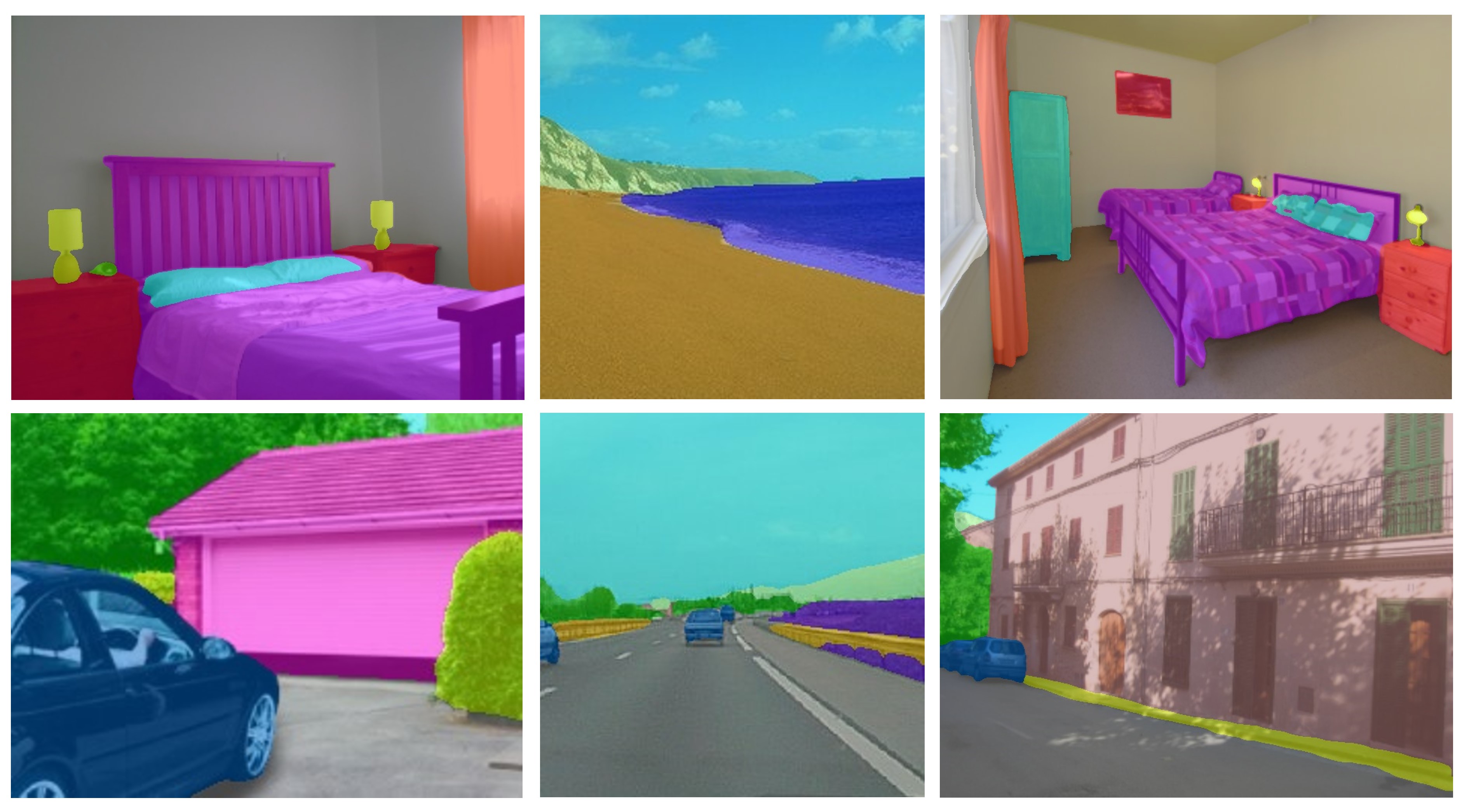} 
\caption{Segmentation results of COSNet on ADE20K validation set. We notice that our framework can effectively segment the objects by using boundary information. }
 \label{fig:ade20k_results}
\vspace{-1.0em}
\end{figure}

We present the qualitative performance of our framework in Fig.~\ref{fig:zero_waste_results}, \ref{fig:spectral_waste_results} and \ref{fig:ade20k_results} on ZeroWaste-f, SpectralWaste and ADE20K datasets respectively. Fig.~\ref{fig:zero_waste_results} shows that the proposed COSNet can effectively segment different waste material types such as soft plastic, rigid plastic, metal, and cardboard  compared to the existing methods including DeepLabv3+ \cite{chen2018_deeplabv3_plus} and FANet \cite{Ali2024FANet}. 
Although, segmentation results of FANet \cite{Ali2024FANet} seems to be promising, however, it introduces more false segmentation as highlighted in the second and third rows of the figure. 
\begin{table}[t]
\centering
\caption{Comparison of COSNet with state-of-the-art methods on ADE20K \cite{Zhou2017ScenePT_ade20k} dataset. We report the results in terms of mIoU metric. Here, \textit{SS} refers to single scale and \textit{MS} refers to multi-scale evaluation. The best results are highlighted in bold.}
\scalebox{0.65}{
\setlength{\tabcolsep}{8pt}
\begin{tabular}{l|c|c|c}
\hline
\rowcolor{Gray}
Method & Params (M)$\downarrow$ & Avg. Latency (ms)$\downarrow$ & mIoU(SS/MS) (\%)$\uparrow$ \\
\hline
Swin-T \cite{liu2021swin} & 60 & 90 & 44.5/45.8 \\
FocalNet-T \cite{yang2022focalnet} & 61 & 88 & 46.5/47.2 \\
ConvNext-T \cite{liu2022convnext} & 60 & \textbf{76} &  46.7/-- \\
FocalAtt-T \cite{Yang2021FocalSF} & 62 & -- & 45.8/47.0  \\
InternImage-T \cite{interimage_cvpr_2023} & 59 & 105 & \textbf{47.9}/48.1  \\
\textbf{COSNet (ours)} & \textbf{37} & 85 & 47.3/\textbf{48.4} \\
\hline
\end{tabular}
}
\vspace{-1.0em}
\label{tab:comparison_of_cosnet_on_ade20k}
\end{table}

Similarly, for the challenging scenarios of SpectralWaste \cite{casao2024spectralwaste} dataset, we show that the proposed COSNet can better segment various types of waste objects in Fig.~\ref{fig:spectral_waste_results} as compared to recently introduced FANet \cite{Ali2024FANet}. In the first and second row, our COSNet detects the trash bag and video tape objects which are not detected by FANet \cite{Ali2024FANet}. 
Likewise, proposed COSNet accurately detects the cardboard objects in the third row of Fig.~\ref{fig:spectral_waste_results} as highlighted in yellow boxes.

While Fig.~\ref{fig:zero_waste_results} and \ref{fig:spectral_waste_results} present the challenging scenarios in cluttered background, Fig.~\ref{fig:ade20k_results} illustrate the diverse scenes of ADE20k dataset having greater number of class object. Nevertheless, our COSNet performs favorably and is able to segment the objects with clear boundaries.

\subsection{Ablation Experiments}
We present the effectiveness of proposed contributions on ZeroWaste-f \cite{bashkirova2022_zerowaste} dataset in Tab.~\ref{tab:ablation_on_zero_waste}. 
In the first row of Tab.~\ref{tab:ablation_on_zero_waste}, MCFS, SM, and BEM modules are absent from the network. As a result, the performance of the network is significantly reduced to 37.48\%.

\begin{table}[t!]
\centering
\caption{Ablation study of enhancement modules in COSNet on ZeroWaste-f dataset. We notice that utilization of the sharpening module (SM) in MCFS and BEM in COSNet provides the best performance highlighted in bold text.}
\scalebox{0.7}{
\setlength{\tabcolsep}{7pt}
\begin{tabular}{l|c|c|c|c|c|c}
\hline
\rowcolor{Gray}
& Exp. No. & MCFS & SM & BEM & mIoU (\%)$\uparrow$ & Pix. Acc. (\%)$\uparrow$ \\
\hline
\multirow{4}{*}{COSNet} & 1 & \tikzxmark & \tikzxmark & \tikzxmark & 37.48 & 88.29 \\
& 2 & $\checkmark$ & \tikzxmark & \tikzxmark & 46.31 & 90.44 \\
& 3 & $\checkmark$ & $\checkmark$ & \tikzxmark & 52.38 & 91.29 \\
& 4 & $\checkmark$ & $\checkmark$ & $\checkmark$ & \textbf{56.67} & \textbf{91.91} \\
\hline
\end{tabular}
}
\vspace{-1.0em}
\label{tab:ablation_on_zero_waste}
\end{table}

\begin{figure*}[t!]
\centering
 \includegraphics[width=1.0\linewidth]{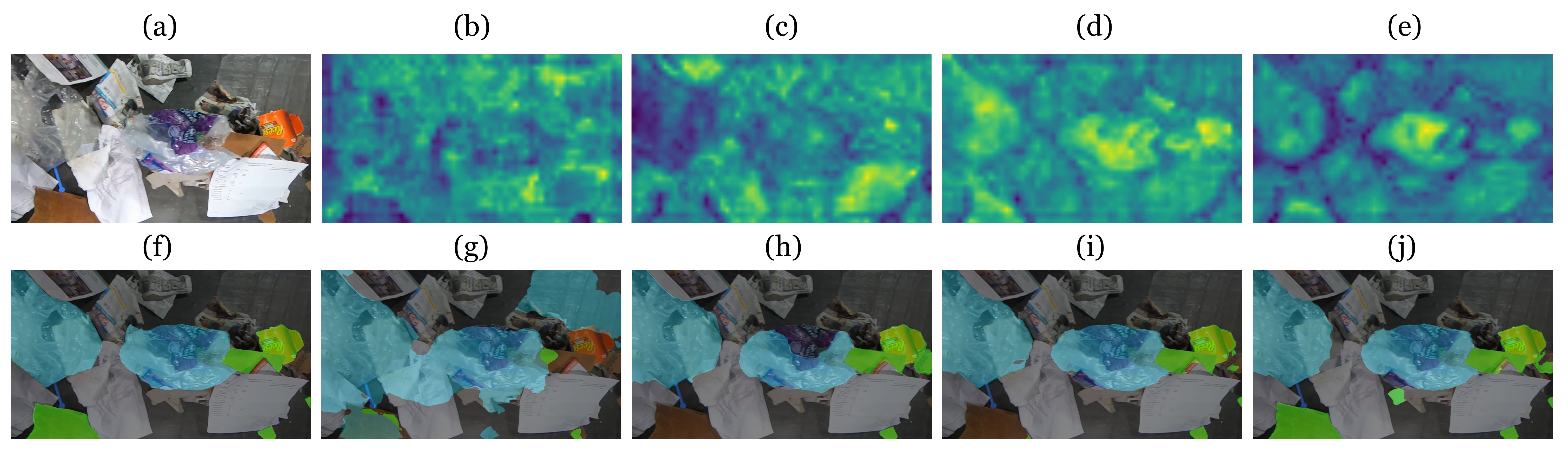} 
\caption{Demonstration of the proposed modules of COSNet on ZeroWaste-f \cite{bashkirova2022_zerowaste} dataset. (a) is the input image, (b)-(e) refer to features of the third stage for the rows (1-4) in Tab. \ref{tab:ablation_on_zero_waste}, respectively. (f) is the ground truth mask, (g)-(j) denote the corresponding prediction masks for rows  (1-4)  in Tab. \ref{tab:ablation_on_zero_waste}, respectively. The above results clearly indicate the efficacy of the proposed modules when integrated in the network.}
\vspace{-0.5em}
 \label{fig:feature_visualization_ablation}
\end{figure*}
\begin{figure*}[t!]
\centering
 \includegraphics[width=1.0\linewidth]{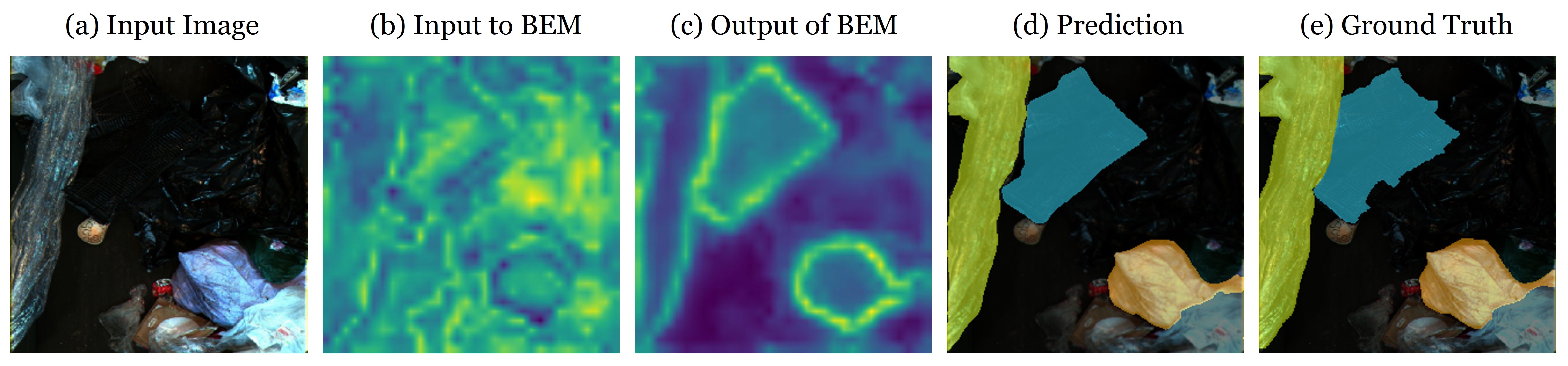} 
\caption{Demonstration of the enhancement capabilities of proposed BEM on SpectralWaste \cite{casao2024spectralwaste} dataset. (a) is the input image to the model, (b) shows the feature map visualization of the third stage of backbone network that is input to BEM, (c) shows enhanced feature maps using BEM highlighting the boundaries of the segmented objects, (d) presents the predicted segmentation mask, and (e) is the ground truth mask.}
\vspace{-1.0em}
 \label{fig:feature_visualization_bem}
\end{figure*}

Adding the MCFS component in the network while removing the SM provides reasonably better mIoU score of 46.31\%. 
Correspondingly, introducing the SM in the MCFS further improves the performance of the network by 6.1\%.
Finally, the addition of the BEM module as an intermediate network in COSNet further enhances the performance and achieves the state-of-the-art result of 56.67\% in terms of mIoU score. It is evident from the Tab.~\ref{tab:ablation_on_zero_waste} that boundary information improves the performance of the COSNet for the segmentation task which validates the efficacy of the proposed contributions.

In addition to the quantitative results reported in Tab. \ref{tab:ablation_on_zero_waste}, we demonstrate the effectiveness of the proposed modules in Fig. \ref{fig:feature_visualization_ablation} and \ref{fig:feature_visualization_bem} respectively. 
It is evident from the Fig. \ref{fig:feature_visualization_ablation} that the introduction of MCFS, SM, and BEM modules in the network improves the segmentation capabilities of the COSNet. 
Similarly, as depicted in Fig. \ref{fig:feature_visualization_bem}, the BEM highlights the boundary information and enhances the features of the third stage by suppressing the unnecessary information. To summarize, COSNet effectively utilize the boundary cues for better segmentation of waste objects in cluttered scenarios.

\section{Conclusion}
In this work, we introduce a semantic segmentation framework named COSNet that utilizes the multi-contextual representations and boundary cues to improve the segmentation performance  in cluttered scenes. COSNet utilizes an enhanced backbone to highlight the boundary information and capture rich feature representations. Additionally, intermediate network comprising of BEM further enhances the object boundaries and suppress the irrelevant information for better segmentation results. Extensive experiments on three challenging datasets reveal the merits of the proposed framework. The potential future direction is to extend the work for segmentation of transparent and camouflaged objects in real-time scenarios.
{\small
\bibliographystyle{ieee_fullname}
\bibliography{egbib}
}

\end{document}